\pdfoutput=1
\documentclass[11pt]{article}

\usepackage[final]{acl}

\usepackage{times}
\usepackage{latexsym}

\usepackage[T1]{fontenc}
\usepackage[utf8]{inputenc}

\usepackage{microtype}

\usepackage{inconsolata}

\usepackage{graphicx}
\usepackage{amsmath}
\usepackage{algorithm}
\usepackage{algorithmic}

\title{MASTER: A Multi-Agent System with LLM Specialized MCTS}

\author{Bingzheng Gan\textsuperscript{1}, Yufan Zhao\textsuperscript{1}, Tianyi Zhang\textsuperscript{1}, Jing Huang\textsuperscript{1}, Yusu Li\textsuperscript{1}\\
{\bf Shu Xian Teo\textsuperscript{1}, Changwang Zhang\textsuperscript{1}, Wei Shi\textsuperscript{1}}\\
  \textsuperscript{1}Huawei Technologies, Co., Ltd.\\
\texttt{\{gan.bingzheng,zhao.yufan1,zhang.tianyi,huangjing114,liyusu,}\\
\texttt{teo.shu.xian,zhangchangwang,w.shi\}@huawei.com}}

\begin{document}
\maketitle
\begin{abstract}
Large Language Models (LLM) are increasingly being explored for problem-solving tasks. However, their strategic planning capability is often viewed with skepticism. Recent studies have incorporated the Monte Carlo Tree Search (MCTS) algorithm to augment the planning capacity of LLM. Despite its potential, MCTS relies on extensive sampling simulations to approximate the true reward distribution, which leads to two primary issues. Firstly, MCTS is effective for tasks like the Game of Go, where simulation results can yield objective rewards (e.g., 1 for a win and 0 for a loss). However, for tasks such as question answering, the result of a simulation is the answer to the question, which cannot yield an objective reward without the ground truth. Secondly, obtaining statistically significant reward estimations typically requires a sample size exceeding 30 simulations, resulting in excessive token usage and time consumption. To address these challenges, we present the \textbf{M}ulti-\textbf{A}gent \textbf{S}ystem with \textbf{T}actical \textbf{E}xecution and \textbf{R}easoning using LLM Specialized MCTS (\textbf{MASTER}), a novel framework that coordinates agent recruitment and communication through LLM specialized MCTS. This system autonomously adjusts the number of agents based on task complexity and ensures focused communication among them. Comprehensive experiments across various tasks demonstrate the effectiveness of the proposed framework. It achieves 76\% accuracy on HotpotQA and 80\% on WebShop, setting new state-of-the-art performance on these datasets.
\end{abstract}

\section{Introduction}
\label{sec:introduction}
LLM represent a significant milestone in artificial intelligence and are increasingly employed in problem-solving tasks \cite{xi2023rise}. However, their application to complex problems is often limited due to concerns about their planning capabilities. LLM generate text based on "next token probability" conditioned on the context \cite{vaswani2017attention,huang2023survey}, but effective reasoning requires rigorous deduction rooted in the first principles of logic, grounded in data and reality \cite{valmeekam2023planbench}.

To enhance the planning capabilities of LLM, recent studies have incorporated the MCTS algorithm \cite{hao2023reasoning, zhou2024language,wang2024promptagent}. MCTS uses simulations to evaluate the long-term consequences of actions, backpropagating aggregated rewards up the tree. This allows for backtracking to previous states based on estimated future potential, striking a balance between exploitation and exploration. However, the use of MCTS in existing works presents two significant challenges: 

\begin{enumerate}
    \item It relies on objective criteria from the external environment to obtain simulation rewards, which are not always available (e.g., in question-answering tasks where the correctness of an answer cannot be determined without ground truth);
    \item It requires numerous simulations to obtain statistically significant rewards, which can be unsustainable due to the time and token costs.
\end{enumerate}

To tackle the first challenge, certain approach compares the outcomes of simulations with the ground truth to derive objective rewards \cite{zhou2024language} which is flawed because revealing the ground truth during problem-solving is inappropriate. In addressing the second challenge, they limit the number of simulations \cite{hao2023reasoning} or terminate the process once a correct answer is identified \cite{zhou2024language}. Yet, this early termination approach is not feasible if the ground truth remains undisclosed.

These issues highlight a critical step of MCTS: simulation. Due to these limitations, MCTS is constrained to a narrow scope of application and is not fully compatible with LLM. Consequently, we propose an adaptation of MCTS tailored to LLM scenarios. Instead of performing limited simulations with uncertain rewards, we eliminate the simulation process, relying on the LLM’s self-evaluation capabilities to allocate rewards. Additionally, we propose several methods to enhance the objectivity of rewards: 1). An additional step is introduced for the LLM to provide more context before self-evaluation; 2). The LLM’s confidence is incorporated as a weight for the reward to regulate its influence; 3). The backpropagation mechanism is retained, allowing for rewards updates if initially misallocated. While traditional MCTS focuses resources on simulations for an approximate reflection of reality, our approach distributes resources across multiple steps to jointly ensure the accuracy and objectivity of rewards. This is why we named our project \textbf{MASTER}, as it replaces simulation, the core procedure of MCTS, by \textbf{mastering} a series of refined designs.

Another contribution of this paper is the introduction of a novel multi-agent system. Current multi-agent systems feature two prominent frameworks: the first allows agents to be independently created and to share ideas freely. While versatile, this open communication can lead to off-topic discussions due to hallucinations \cite{lin2024rella, hong2023metagpt, xi2023rise, zhang2024controlling}, diverting focus from the main task and depleting the token window length for extended conversation history. The second approach involves human-created agents, where communication is predefined. Although more controlled, this approach lacks code reusability \cite{chu2023survey}. Additionally, since the procedures are established, it cannot adapt to tasks of varying difficulty. It struggles with unanticipated complex tasks on the one hand and spends unnecessary resources on easy tasks on the other.

Our system, MASTER, addresses these limitations by employing LLM-specialized MCTS to guide the creation and interaction of agents. In this system, child agents respond to and build upon the outputs of the parent agent, making recruitment and communication more manageable and efficient. The system dynamically scales the number of agents based on task complexity, ensuring flexibility. Although the agents share similar profiles, they assume different roles by taking distinct actions. Unlike the nodes in LATS \cite{zhou2024language} and RAP \cite{hao2023reasoning}, which represent states on a reasoning tree and are closely tied to specific tasks, our agents are task-agnostic and do not require reconfiguration when the task changes.

\begin{figure*}[ht]
    \centering
    \includegraphics[width=\linewidth]{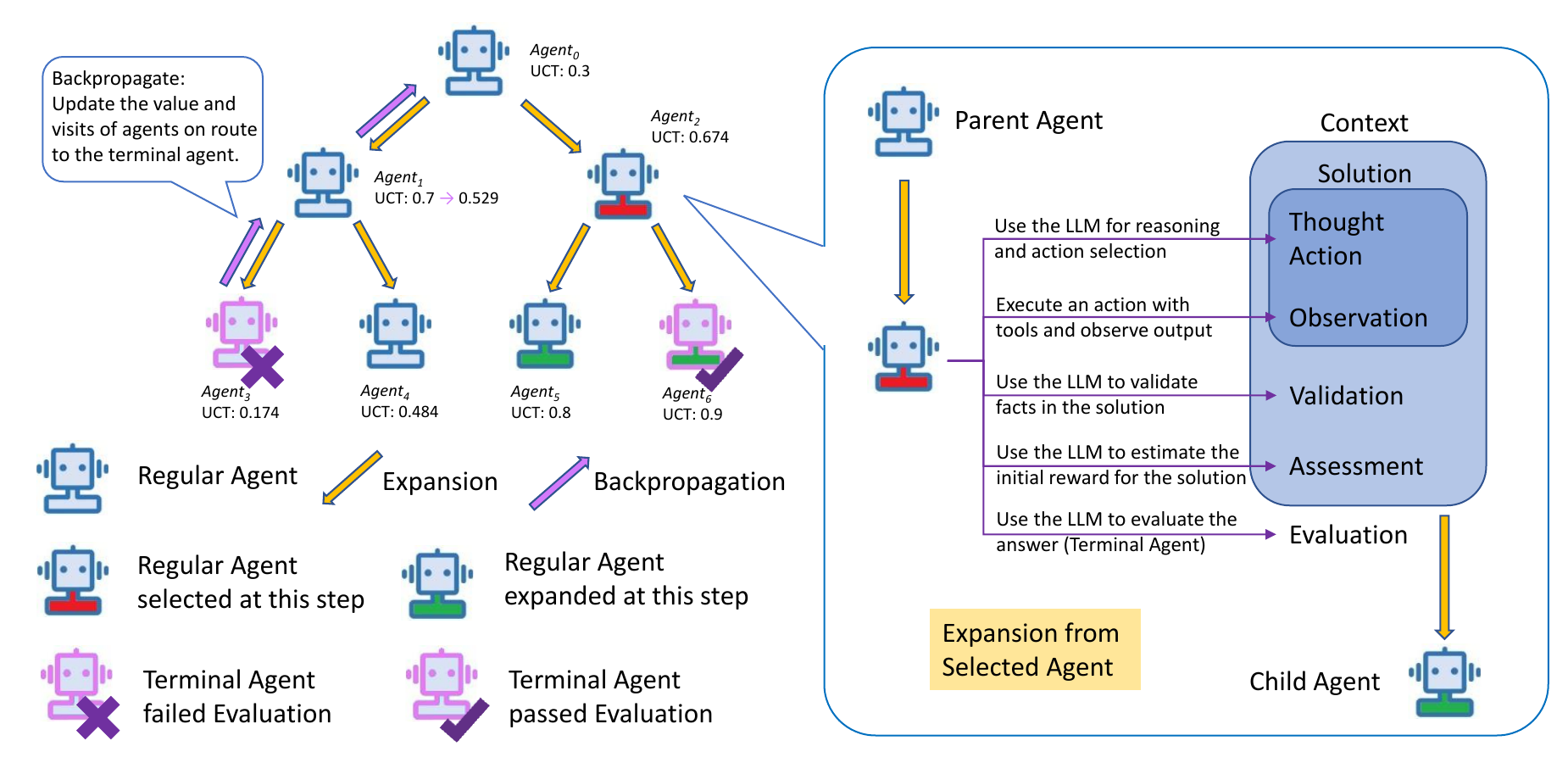}
    \caption{Reasoning Tree of MASTER. Starting from $\text{Agent}_0$, $\text{Agent}_1$ and $\text{Agent}_2$ are created in the first expansion. Then the system first selects $\text{Agent}_1$ for expansion due to its higher UCT. Its child agent $\text{Agent}_3$ is a terminal agent that failed evaluation which triggers a backpropagation and lowers the UCT of $\text{Agent}_1$. Now $\text{Agent}_2$ has the highest UCT and is selected for next expansion. Its child agent, $\text{Agent}_6$ is a terminal agent and passes evaluation. The answer in it is the final answer.}
    \label{fig:final_figure}
\end{figure*}

In summary, our key contributions are:
\begin{enumerate}
    \item We propose a novel \textbf{M}ulti-\textbf{A}gent \textbf{S}ystem with \textbf{T}actical \textbf{E}xecution and \textbf{R}easoning using LLM Specialized MCTS (\textbf{MASTER}), a novel multi-agent framework that employs a new agent recruitment process and communication protocol based on the MCTS algorithm. The system autonomously adjusts the number of agents according to task complexity and mitigates distractions and shortage of token window during agent communication.
    \item We introduce a modified version of MCTS tailored to LLM. This adaptation is suitable for tasks where the environment does not provide objective feedback, addressing a limitation of the original MCTS. This revised MCTS is implemented within our MASTER framework.
    \item We conduct comprehensive experiments accross diverse tasks, including Question Answering (HotpotQA), Decision Making (WebShop), and Programming (MBPP). It achieves 76\% accuracy on HotpotQA and 80\% on WebShop, setting new SOTA on these datasets.
\end{enumerate}

\section{Related Work}
\label{sec:relatedwork}
Many studies have been done to enhance the planning capabilities of LLM. Among these efforts, two primary planning approaches have emerged for agents: Single-Path Planning and Tree-Based Planning. In the context of multi-agent systems, current frameworks predominantly employ either Predefined Frameworks or Open Frameworks depending on their agent communication patterns. We discuss some of the relevant works in this section.

\subsection{Planning Processes}
\label{subsec:planningprocesses}
\subsubsection{Single-Path Planning}
\label{subsubsec:singlepathplanning}
In single-path planning, the LLM follows one trajectory at a time, without branching into multiple possibilities. Early examples include Few-Shot Prompting \cite{brown2020language}, where the LLM is guided by examples of completed tasks, and Chain of Thought approaches \cite{wei2022chain,kojima2023large,ning2024skeletonofthought}, which require the LLM to reason step-by-step, maintaining a linear trajectory throughout the process. Zhang et al. introduce a structured meta-prompt with placeholders for the LLM to complete \cite{zhang2024meta} while Suzgun and Kala provide task-related information to guide the model's path \cite{suzgun2024metaprompting}. Single-path planning also benefits from external feedback to refine solutions. ReAct \cite{yao2023react} integrates feedback from the environment, and Reflexion \cite{shinn2023reflexion} supplements this with verbal reasoning based on the received feedback. Chen et al. use outputs and error messages from a code interpreter to assist the LLM in debugging \cite{chen2024teaching}, while Qiu et al. leverage outputs from a symbolic interpreter to enhance the LLM’s inductive reasoning capabilities \cite{qiu2024phenomenal}.

\subsubsection{Tree-Based Planning}
\label{subsubsec:treebasedplanning}
In complex problem-solving, it is often beneficial to explore multiple thought trajectories and backtrack as needed. Tree-based planning organizes these thoughts into a tree structure, on which a search algorithm is applied. For example, BFS/DFS is employed in the Tree of Thoughts \cite{yao2023tree}. RAP \cite{hao2023reasoning} and LATS \cite{zhou2024language} utilize MCTS to approximate reality and support the LLM’s reasoning processes. However, how much the simulation process contributes to their success remains uncertain due to those two issues mentioned in Introduction section. To address the first challenge, RAP prompts the LLM with "Is this reasoning step correct?" and uses the next-word probability of "yes" as a reward, thus leveraging the LLM's evaluation capabilities without relying on external criteria. For the second challenge, RAP reduces costs by performing only one simulation. From a mathematical perspective, it is questionable that one simulation can accurately approximate the real reward. Some methods stop simulations once a correct answer is found. However, this early termination mechanism is unavailable if the ground truth is not revealed.

\subsection{Multi-Agent Systems}
\label{subsec:multiagentsystems}
\subsubsection{Predefined Framework}
\label{subsubsec:predefinedframework}
In predefined frameworks, the recruitment and communication of agents are structured in advance. For instance, ChatDev \cite{qian2023communicative} and MetaGPT \cite{hong2023metagpt}, both tailored for software development, rely on predefined workflows. Similarly, AutoAgents \cite{chen2024autoagents}, a framework designed for automatic agent generation, follows a predefined structure. These frameworks have been criticized for their heavy dependence on upfront planning and their lack of flexibility in responding to changing requirements \cite{Pargaonkar2023Waterfall}.

\subsubsection{Open Framework}
\label{subsubsec:openframework}
Conversely, open frameworks offer greater flexibility, allowing agents to interact more dynamically. For example, AgentVerse \cite{chen2024agentverse} recruits agents with the freedom to communicate openly, while CAMEL \cite{li2023camel} explores a two-agent system. Additionally, AutoGen \cite{wu2023autogen} facilitates next-generation LLM applications by enabling multi-agent conversations, allowing for adaptive and context-aware interactions. As noted by Wei et al., effective multi-agent collaboration in these open frameworks requires autonomous systems to regulate communication, ensuring that agents know when and with whom to interact to avoid the exchange of irrelevant information which is challenging for these frameworks \cite{wei2023multiparty}.

\section{Methodology}

\label{sec:methodology}
\subsection{Preliminaries}
\label{subsec:mcts}
Before introducing our framework, we present MCTS to clarify the motivation behind our work. MCTS \cite{coulom2006mcts} is a widely used planning algorithm and famously employed in AlphaGo \cite{alphago}. Taking the Game of Go as an example, the algorithm assists in selecting the best possible action in the current state of the board based on their average rewards. These rewards are obtained through numerous simulations. For instance, consider the current state where action $a_1$ is chosen as the next move. The game then proceeds to completion, with all subsequent actions, whether by our side or the opponent, determined by a policy model rather than a real player. The entire game sequence constitutes one simulation of action $a_1$. If we win, the reward is 1; otherwise, it is 0. Specifically, if we simulate 10 games from the current state with action $a_1$ and win 9 of them, the average reward for $a_1$ would be 0.9. However, due to the vast action space in the Game of Go, it is impractical to simulate every possible action. The Upper Confidence Bound 1 applied to Trees (UCT) identifies actions with higher potential to win, allocating more simulations to these actions rather than distributing simulations equally among all actions. Once an action is decided based on this process and physically executed, leading to a new game state, the same procedure is then applied to select actions from this new state, and the planning continues until the actual end of the Game of Go.

MCTS typically involves four key procedures: \textbf{Selection:} Traverse the current reasoning tree to select the node with the highest UCT for simulation. \textbf{Expansion:} Extend the reasoning tree by adding new child nodes from the selected node. \textbf{Simulation:} Continue expanding from a child node until the task is completed. \textbf{Backpropagation:} After each simulation, update the average reward of corresponding node with the newly obtained simulation reward.

\subsection{Framework of MASTER}
\label{subsec:framework}
In our framework, MASTER, the concepts of reasoning tree, selection, expansion, and backpropagation are analogous to those used in MCTS. However, we eliminate the simulation step due to those two issues highlighted in the Introduction section. Instead, our framework incorporates three special mechanisms to derive rewards: providing more context before self-evaluation, incorporating LLM’s confidence in our novel UCT formula, and updating rewards in backpropagation. The details of our framework are demonstrated in Figure \ref{fig:final_figure} and further explained in following paragraphs.

When our framework is presented with a task, it initializes a root agent with the problem. This agent then prompts the LLM to generate text that reflects on the current reasoning trace (\textbf{Thought}) and proposes an action to solve the problem (\textbf{Action}). The model uses a temperature of 0.6 to encourage more diverse thoughts and actions in the reasoning tree. The agent then executes this action using tools that interact with the external environment, producing feedback (\textbf{Observation}). The texts, including Thought, Action, and Observation, form the agent’s \textbf{Solution}, as illustrated in Figure \ref{fig:final_figure}. Subsequently, the agent prompts the LLM to generate a textual output that verifies key facts in the current Solution (\textbf{Validation}). Finally, based on the Solution and Validation, the agent prompts the LLM to evaluate progress toward solving the problem, generating both a score and the LLM’s confidence in this score (\textbf{Assessment}). These two values (score and confidence) are then extracted. The model uses a temperature of 0.0 for Validation and Assessment to ensure more stable decisions. All texts (Solution, Validation, and Assessment) form the agent’s \textbf{Context} and are stored in its memory (Figure \ref{fig:final_figure}). The score serves as the agent’s initial reward.

The previous paragraph describes the generation of a single agent in our system. After the root agent is generated, child agents are created following the same procedure, with the root agent's Context appended to the prompts of these child agents, as they need to continue solving the problem. As illustrated in Figure \ref{fig:final_figure}, two child agents ($\text{Agent}_1$ and $\text{Agent}_2$) are generated using exactly the same procedure and prompt to explore diverse reasoning paths from the same state (parent agent). The number of child agents is a hyperparameter, \textit{Number of Branches}, which varies depending on the task. The creation of these child agents is termed \textbf{Expansion} from the parent agent.

Further expansion can be carried out from any existing agent using the same procedure. The UCT of each agent is calculated, and the agent with the highest UCT is selected for further expansion. Another hyperparameter, \textit{Maximum of Expansion}, represents the approximate number of steps required to solve the problem, allowing users to set it based on their understanding of the task. If this limit is reached without finding a satisfactory solution, the solution from the terminal agent with the highest reward is submitted as the final answer.

During the expansion of the reasoning tree, agents that generate a final answer rather than an intermediate step in their Solution, are called \textbf{Terminal Agents}. For instance, in the HotpotQA task, if an agent's Action is 'Finish[]', it is identified as a terminal agent, as this Action indicates a final answer. Similar indicators exist in the Solution of other tasks. During the \textbf{Evaluation} (Figure \ref{fig:final_figure}) which applies only to Terminal Agents, the LLM assesses the correctness of the Solution. If the solution is deemed correct, it is submitted as the final answer, concluding the task. If not, \textbf{Backpropagation} is triggered, using the reward from this terminal agent to update the rewards of all agents on the path up to the root agent. Pseudo-code can be found in Appendix \ref{sec:appendix_a}.

\subsection{Formula of Modified UCT}
\label{subsec:formula}
Recent works RAP and LATS directly apply the original UCT formula. To better suit our design, we propose a modified UCT formula.

The original UCT formula is derived from Hoeffding's Inequality \cite{inbookcite} which can be found in Appendix \ref{sec:appendix_b}. It is typically applied in the following scenario: Given a node representing a state (referred to as $\text{node}_h$), there are multiple subsequent actions to choose from (e.g., $a_i, a_j, a_k$). To determine the Q value of these actions, multiple simulations are conducted, and UCT is employed to decide which action should be simulated. Instead of simply selecting the node with the highest Q value (pure exploitation, the first term in Eqn \ref{eq:methodology1}), UCT balances exploitation and exploration by incorporating an exploration term (the second term in Eqn \ref{eq:methodology1}) that favors nodes with fewer simulations. The $\text{node}_i$, representing the state resulting from action $a_i$, is one of the child nodes of $\text{node}_h$. The UCT formula for $\text{node}_i$ is:

\begin{equation}
  \label{eq:methodology1}
  UCT=Q_i+\sqrt{\dfrac{\ln{\left(N_i\right)}}{2n_i}}
\end{equation}

\begin{equation}
  \label{eq:methodology2}
  Q_i= \dfrac{\sum_{n=1}^{n_i} r_n}{n_i}
\end{equation}

Where 
$n_i$ is the number of backpropagations applied to the $\text{node}_i$.
$r_n$ is the reward of the n-th backpropagation.
$Q_i$ is the estimation of Q value calculated by Eqn \ref{eq:methodology2}. It represents the average reward from simulations.
$N_i$ is the total number of simulations by the parent node of the current $\text{node}_i$ which is $\text{node}_h$. In other words, $N_i$ is the sum of $n_i$, $n_j$ and $n_k$ in above example although $n_j$ and $n_k$ are not explicitly shown in the UCT formula.

In our system, $Q_i$, as the estimation of Q value, consists of two components: \textbf{Initial Reward} is extracted from Assessment when this agent is generated. \textbf{Updating Reward} is the mean of rewards from backpropagation, similar to $Q_i$ in Eqn \ref{eq:methodology2}. Inspired by the research on auxiliary information about rewards in the form of control variables \cite{verma2021stochastic}, we modify the reward estimation in our system as shown in Eqn \ref{eq:methodology3}:

\begin{equation}
  \label{eq:methodology3}
  Q_i=c_{0} \cdot r_{0}+\left(1-c_{0}\right) \cdot \dfrac{\sum_{n=1}^{n_i} r_n}{n_i}
\end{equation}

Where 
$r_{0}$ is the initial reward given by LLM.
$c_{0}$ is the confidence of the LLM to this initial reward.
$r_n$ and $n_i$ are the same with Eqn \ref{eq:methodology2}.

The original MCTS relies heavily on numerous simulations to make this estimation accurate. Consequently, it becomes unreliable when the number of $n_i$ is small. On the other hand, our Q value has an extra component (initial reward), as our Q value is the weighted sum of initial reward and updating reward with the confidence as weight. When LLM has high confidence to the initial reward it assigns, the influence of rewards from backpropagation (updating reward) is reduced due to its lower weight $\left(1-c_{0}\right)$. Conversely, when the LLM has low confidence, updating reward is required to dominate the Q value estimation while the weight of updating reward $\left(1-c_{0}\right)$ is higher. Notably, the number of backpropagation $n_i$ is automatically adapted to each question rather than being manually set by users. For complex tasks, the model requires more attempts to obtain an acceptable answer, with each failed attempt triggering a backpropagation. For simple question, the model may produce an acceptable result in their first attempt, eliminating the need for backpropagation. This early termination mechanism completes the task while reducing token consumption.

In the formula used by RAP and LATS, an exploration constant $\lambda$ replaces the fixed $1/\sqrt{2}$ as the weight of the exploration term as following:

\begin{equation}
  \label{eq:methodology4}
  UCT=Q_i+\lambda \cdot \sqrt{\dfrac{\ln{\left(N_i\right)}}{n_i}}
\end{equation}

In our approach, we use $1 / \left(10 \sqrt{2} c_{0}\right)$ as the exploration weight. The significance of this adjustment is twofold: 1). The exploration term reflects the uncertainty associated with this agent. When the LLM has low confidence in the initial reward, the agent's uncertainty is relatively high, necessitating more exploration. In such cases, a higher exploration weight increases the UCT, guiding the algorithm to select this agent for further exploration; 2).The number of backpropagations in our system is significantly lower than in the original MCTS while the slope of the logarithmic function is relatively steep when the variable is small, so the value of the exploration term tends to play a dominant role. Therefore, this exploration weight should be used to control the influence of this term. Moreover, when the minimum confidence of 0.1 is used, the exploration weight equals $1/\sqrt2$, the same as the weight in Eqn \ref{eq:methodology1}.

%\begin{align}
%  \label{eq:methodology5}
%  UCT = c_{init} \cdot r_{init}+\left(1-c_{init}\right) \cdot \frac{\sum_{n=1}^{n_i} r_n}{n_i} \nonumber \\ +\frac{1}{10 \sqrt{2} c_{init}} \cdot \sqrt{\frac{\ln{\left(N_i\right)}}{n_i}}
%%\end{align}
In summary, the revised formula in our system is:
\begin{equation}
  \label{eq:methodology5}
  UCT = 
  \begin{cases}
    r_{0} & \text{if } n_i = 0 \\
    c_{0} \! \cdot \! r_{0} \! + \! \left(1 - c_{0}\right) \! \cdot \! \dfrac{\sum\limits_{n=1}^{n_i} r_n}{n_i} \\
    + \dfrac{1}{10 \sqrt{2} c_{0}} \cdot \sqrt{\dfrac{\ln{N_i}}{n_i}} & \text{otherwise}
  \end{cases}
\end{equation}
When $n_i$ is 0, indicating that no backpropagation has been applied to the node, the UCT of this node is set to be its initial reward $r_0$.

\subsection{Strategies of Reward Assignment}
\label{sec:strategy}
To conclude, the three special mechanisms we implement to ensure the reliability of rewards in our framework are:

\begin{enumerate}
    \item Before assigning a reward in the Assessment phase, an additional Validation step is conducted, where the LLM comments on the correctness of the facts in the current solution. These comments are added to the prompt for the Assessment step, guiding the LLM toward a more reliable reward. This design is motivated by the observation that the LLM performs better when it is asked to address one problem at a time. Separately verifying correctness and progress can lead to stable and reliable scoring.
    \item In the Assessment phase, the LLM is asked to provide both a score and its confidence in that score, rather than just the score alone. The confidence value plays two roles in our modified UCT formula, which is detailed in the Formula of Modified UCT subsection. If the LLM has low confidence in the score it provides, the influence of this score is reduced and the likelihood to select this agent for further exploration is increased.
    \item Backpropagation occurs after each simulation in the original MCTS. Although we have removed simulations, backpropagation is retained in our framework. It is triggered whenever a terminal agent produces a Solution that fails Evaluation. A failed Evaluation indicates that the reasoning steps leading to this terminal agent may be flawed, and their Q value should be reduced accordingly. This mechanism allows for the adjustment of rewards if they are inaccurate at the outset.
\end{enumerate}

\section{Experiment Setup}
\label{sec:experiments}
To demonstrate the generalizability of our framework, we conduct experiments across a diverse set of tasks, including question answering (HotpotQA), decision making (WebShop), and coding (MBPP). These datasets are widely recognized benchmarks in their respective domains.

In addition to assessing effectiveness and efficiency, we also performed ablation and parameter studies to investigate the contributions of each mechanism in our framework and the impact of hyperparameters.

\subsection{Datasets}
\label{subsec:datasets}
\textbf{HotpotQA}: \cite{yang2018hotpotqa} tests multi-hop reasoning in LLM, requiring models to parse and reason across multiple paragraphs. We used the Distractor setting, where the task is to answer the question with a mix of relevant and irrelevant paragraphs context. \textbf{WebShop}: \cite{NEURIPS2022_82ad13ec} simulates an e-commerce environment to test decision-making abilities. The task involves navigating a virtual store to find products that best match a given instruction, with success measured by how closely the selected product matches the requirement. \textbf{MBPP}: \cite{austin2021program}) assesses coding abilities. Each task includes a problem description and test cases for validation. In our system, agents generate and test complete code, with child agents iteratively improving on errors identified by their parent agents. A task is considered solved when the generated code passes all test cases.

\subsection{Baselines}
\label{subsec:baselines}
Since we use GPT-4, a highly capable LLM, as the base model, we take \textbf{GPT-4} itself without any agent, as a baseline. It is prompted with the task description and the same examples as our framework to solve tasks in a single call (\textbf{Few-shot Chain-of-Thought}). Notably, in this setting, GPT-4 cannot solve HotpotQA and WebShop problems because these tasks require multiple interactions with the environment. For HotpotQA, the model must generate search keywords, receive the retrieved context from the environment, and decide whether to search for additional context or answer the question. For WebShop, the model must purchase a target item on a mock website through multiple search and click actions. Expecting GPT-4 to perform these actions without environment feedback in a single call is impractical. When incorporating external environment feedback to enable multi-turn interactions, the setup becomes identical to the experimental conditions of ReAct. In other words, ReAct's performance in Table \ref{tab:effectiveness} serves as an indicator of the base model's capacity in these two tasks.

\textbf{ReAct} and \textbf{Reflexion} are well-known methods in the planning domain, and our work integrates some of their ideas. \textbf{LATS}, like our approach, is a tree-based method and demonstrates strong performance. Therefore, we compare against these baselines across all three datasets. \textbf{MetaGPT} and \textbf{AgentVerse}, two representative multi-agent systems, serve as our benchmarks in the multi-agent setting. We evaluate them only on MBPP because MetaGPT is specifically designed for programming tasks, while AgentVerse requires execution tools, and only the tool for programming tasks is available.

Additionally, we benchmark our framework against the current state-of-the-art (SOTA) for each dataset: \textbf{Beam Retrieval} for HotpotQA, \textbf{AgentKit} for WebShop, and \textbf{AgentCoder} for MBPP. However, AgentKit is evaluated on two datasets in its original paper (Crafter and WebShop), while its GitHub repository provides code only for Crafter. Moreover, the available code for Crafter cannot be adapted to WebShop due to insufficient implementation details. Consequently, we rely on the original performance claims from their paper. Nevertheless, given the substantial performance gap between our results and theirs, we believe our SOTA claims remain highly plausible.

\subsection{Implementation Details}
\label{subsec:implementation}
Given the high cost of GPT-4, we randomly select a sample of 100 questions from each dataset, following the approach used in Reflexion \cite{shinn2023reflexion} and LATS \cite{zhou2024language}. To ensure fairness, the same random seed is used across all three datasets to select these 100 questions.

To mitigate the effect of LLM randomness on accuracy, we repeat each experiment three times on the same samples and report the mean accuracy in Table \ref{tab:effectiveness}. Our framework effectively controls LLM randomness through the strategies outlined in the Methodology section, where multiple steps support and verify each other.

\section{Results and Analysis}
\label{sec:results}
\subsection{Effectiveness Analysis}
\label{subsec:effectiveness}
We reproduce all baseline approaches, except for AgentKit due to insufficient information, with GPT-4 as base model on the same 100 questions and record the results. Additionally, we compare these results with those reported in the respective papers. The better result is used as the final value for each baseline in Table \ref{tab:effectiveness}. This method favors baseline approaches and ensures that our framework demonstrates superior performance across various standards.

\begin{table}[]
    \resizebox{\columnwidth}{!}{%
    \begin{tabular}{llll}
                                        & HotpotQA & WebShop  & MBPP     \\ \hline
    \multicolumn{1}{l|}{GPT-4 (CoT)}    & -        & -        & 0.683    \\
    \multicolumn{1}{l|}{ReAct}          & 0.420    & 0.320    & 0.710    \\
    \multicolumn{1}{l|}{Reflexion}      & 0.510    & 0.350    & 0.771    \\
    \multicolumn{1}{l|}{LATS}           & 0.710    & 0.380    & 0.811    \\
    \multicolumn{1}{l|}{MetaGPT}        & -        & -        & 0.877    \\
    \multicolumn{1}{l|}{AgentVerse}     & -        & -        & 0.890    \\
    \multicolumn{1}{l|}{Beam Retrieval} & 0.733    & -        & -        \\
    \multicolumn{1}{l|}{AgentKit}       & -        & 0.702    & -        \\
    \multicolumn{1}{l|}{AgentCoder}     & -        & -        & 0.918    \\ \hline
    \multicolumn{1}{l|}{Ours}           & 0.760    & 0.800    & 0.910    \\ \hline   
    \end{tabular}%
    }
    \caption{Effectiveness comparison with accuracy.}
    \label{tab:effectiveness}
\end{table}

MASTER sets new SOTA performance across multiple tasks: 1). 76.0\% Exact Match accuracy on HotpotQA, surpassing Beam Retrieval’s 73.3\% \cite{zhang2024endtoend}; 2). 80.0\% accuracy on WebShop, exceeding AgentKit’s 70.2\% \cite{wu2024agentkit}.
On MBPP, it closely matches AgentCoder’s 91.8\% pass@1 accuracy with 91.0\% \cite{huang2024agentcoder}, showing competitive performance in programming task.

\subsection{Efficiency Analysis}
\label{subsec:efficiency}
As a tree-based method, token consumption is a concern due to the diverse reasoning trajectories. However, by removing the simulation step and introducing an early termination mechanism, our framework achieves higher efficiency compared to other tree-based methods using MCTS. We benchmark our efficiency against LATS because: 1). LATS is a typical tree-based approach with superior performance compared to similar frameworks; 2). LATS reports the lowest token consumption when compared with ToT \cite{yao2023tree} and RAP \cite{hao2023reasoning} in their paper.

We measure token consumption for LATS (n = 5, k = 50) and MASTER (\textit{number of branches} = 2, \textit{maximum of expansion} = 3) on the same 100 HotpotQA questions, using the same hyperparameter settings as those in the effectiveness analysis experiments. The average cost per question was 185,392 tokens for LATS and 10,937 tokens for MASTER. Our approach uses only about 6\% of the tokens compared to LATS while delivering better performance (Table \ref{tab:effectiveness}).

LATS \cite{zhou2024language} reports an average cost of 173,290 tokens per question, lower than our reproduced results. This discrepancy may be due to the fact that their tests were conducted on correctly answered questions, while our tests included some incorrectly answered questions, which tend to consume more tokens as the algorithm continues until reaching the maximum trial limit.

\subsection{Ablation Study}
\label{subsec:ablationstudy}
\subsubsection{UCT Modification}
In MCTS, the UCT formula balances exploration and exploitation. One of our main contributions is adapting this formula to better accommodate LLM. We evaluate the impact of removing components of our modified UCT formula, considering the following cases (Table \ref{tab:ablation-uct}): 
\begin{enumerate}
    \item Our full modified formula, incorporating the weighted sum of initial and updating rewards, and exploration weight influenced by the LLM's confidence (Eqn \ref{eq:methodology5}).
    \item A variant with the weighted sum of rewards and a fixed exploration weight (Eqn \ref{eq:experiment1}). This differs from the full formula by not incorporating the LLM's confidence in the exploration term.
\begin{align}
  \label{eq:experiment1}
  UCT = c_{0} \cdot r_{0}+\left(1-c_{0}\right) \cdot \dfrac{\sum_{n=1}^{n_i} r_n}{n_i} \nonumber \\ + \sqrt{\dfrac{\ln{\left(N_i\right)}}{n_i}}
\end{align}
    \item A variant using only the weighted sum of rewards (Eqn \ref{eq:experiment2}), removing the entire exploration term.
\begin{equation}
  \label{eq:experiment2}
  UCT=c_{0} \cdot r_{0}+\left(1-c_{0}\right) \cdot \dfrac{\sum_{n=1}^{n_i} r_n}{n_i}
\end{equation}
    \item A variant using only the initial reward for exploitation (Eqn \ref{eq:experiment3}), excluding exploration term and updating reward from backpropagation.
\begin{equation}
\label{eq:experiment3}
    UCT=r_{0}
\end{equation}
\end{enumerate}

When using UCT with a fixed exploration weight (case 2), performance declines on two out of three datasets, even worse than the variant without the entire exploration term. As discussed in the Methodology section, the exploration term plays a dominant role and should be modulated by the exploration weight. This result supports that hypothesis, as the system sometimes over-explores without progressing toward task completion when using this formula. Omitting a dynamic exploration weight based on confidence is harmful to the system.

Using the weighted sum of initial and updating rewards (case 3) performs better than using the initial reward alone (case 4), likely because the updating reward incorporates additional information from deeper in the reasoning tree.

The MBPP results indicate that different UCT variants have no significant effect on this dataset. This outcome likely arises because the Observation, represented by the test case results, is objective. The Observation is appended to the prompt of the Assessment, allowing the LLM to assign rewards with high confidence. When $c_{0}$ is 1, all UCT variants reduce to case 4 or very close to it. Therefore, all settings yield identical or nearly identical outcomes.

\begin{table}[t]
    \resizebox{\columnwidth}{!}{%
    \begin{tabular}{llll}
                                                  & HotpotQA       & WebShop        & MBPP           \\ \hline
    \multicolumn{1}{l|}{Full modified UCT}        & \textit{0.760} & \textit{0.800} & \textit{0.910} \\
    \multicolumn{1}{l|}{Fixed Exploration Weight} & 0.700          & 0.677          & 0.910          \\
    \multicolumn{1}{l|}{w/o Exploration Term}     & 0.737          & 0.750          & 0.910          \\
    \multicolumn{1}{l|}{Only Initial Reward}      & 0.723          & 0.703          & 0.910          \\ \hline        
    \end{tabular}%
    }
    \caption{Ablation study with removing different components of our modified UCT formula.}
    \label{tab:ablation-uct}
\end{table}

\begin{table}[t]
    \resizebox{\columnwidth}{!}{%
    \begin{tabular}{llll}
                                            & HotpotQA       & WebShop        & MBPP           \\ \hline
    \multicolumn{1}{l|}{Full setting}       & \textit{0.760} & \textit{0.800} & \textit{0.910} \\
    \multicolumn{1}{l|}{w/o Validation}     & 0.623          & 0.563          & 0.863          \\
    \multicolumn{1}{l|}{w/o Assessment}     & 0.233          & 0.157          & 0.743          \\ \hline        
    \end{tabular}%
    }
    \caption{Ablation study with removing validation or assessment of our agent.}
    \label{tab:ablation-agent}
\end{table}

\begin{table}[t]
    \resizebox{\columnwidth}{!}{%
    \begin{tabular}{llll}
    \multicolumn{1}{l|}{Number of Branches}     & HotpotQA       & WebShop        & MBPP           \\ \hline
    \multicolumn{1}{l|}{1}                      & 0.733          & 0.797          & 0.903          \\
    \multicolumn{1}{l|}{2 (All datasets used)}  & \textit{0.760} & \textit{0.800} & \textit{0.910} \\
    \multicolumn{1}{l|}{3}                      & 0.763          & 0.797          & 0.910          \\ \hline
    \end{tabular}%
    }
    \caption{Parameter study on Number of Branches.}
    \label{tab:parameter-branch}
\end{table}

\begin{table}[t]
    \resizebox{\columnwidth}{!}{%
    \begin{tabular}{llll}
    \multicolumn{1}{l|}{Maximum of Expansion}     & HotpotQA       & WebShop        & MBPP           \\ \hline
    \multicolumn{1}{l|}{1}                        & 0.000          & 0.000          & 0.747          \\
    \multicolumn{1}{l|}{3 (HotpotQA \& MBPP)}     & \textit{0.760} & 0.013          & \textit{0.910} \\
    \multicolumn{1}{l|}{8 (WebShop)}              & 0.770          & \textit{0.800} & 0.910          \\ \hline
    \end{tabular}%
    }
    \caption{Parameter study on Maximum of Expansion.}
    \label{tab:parameter-expansion}
\end{table}

\subsubsection{Agent Design}
The validation step before assessment is another key feature of our framework. We conduct additional ablation studies, removing the validation and assessment steps individually. Since the initial reward is derived from the assessment and the algorithm cannot function without it, we assign random initial rewards when the assessment step is removed (Table \ref{tab:ablation-agent}).

Performance drops significantly when the validation or assessment step is removed, even for the MBPP dataset, as validation greatly impacts reward allocation, the core component of the MCTS. Additionally, with random rewards assigned when the assessment step is removed, performance essentially relies on the LLM alone, or worse.

\subsection{Parameter Study}
\label{subsec:parameterstudy}
We conduct parameter studies to determine optimal values for two key hyperparameters: \textit{Number of Branches} (Table \ref{tab:parameter-branch}) and \textit{Maximum of Expansion} (Table \ref{tab:parameter-expansion}), across all three datasets.

The \textit{Number of Branches} has little effect on performance in WebShop and MBPP. However, on HotpotQA, performance drops by nearly 3\% when reducing it from 2 to 1. This reduction may hinder the system, as multiple reasoning trajectories help prevent getting stuck in incorrect states. Although WebShop faces similar challenges, the agent can use the dataset in-built action 'prev' to mitigate, though not completely avoid, this problem. To balance performance and cost, we use 2 for all datasets.

We select 1, 3, 8 for the \textit{Maximum of Expansion} here because in our other experiments, 3 is used in HotpotQA and MBPP while 8 is used for WebShop. Normally, questions in HotpotQA and WebShop require multiple steps to solve so their performance drops dramatically when the \textit{Maximum of Expansion} is lower than the steps needed to solve the problem, as they are forced to stop before getting an answer.

\section{Conclusion}
\label{sec:conclusion}
This paper introduces MASTER, a novel multi-agent system framework that leverages a specialized MCTS to enhance the planning capabilities of LLM. Our LLM-optimized MCTS broadens the applicability of MCTS to a wider range of tasks with reduced costs. Besides, we employ this algorithm to guide agent recruitment and communication protocols, thereby introducing an innovative form of multi-agent system. Extensive experiments across various datasets demonstrate MASTER's effectiveness and efficiency over existing frameworks.

\section{Limitations}
\label{sec:limitations}
There are some limitations in our framework. Firstly, it relies heavily on the LLM's ability to provide accurate scores and confidence assessments of the current reasoning state. While GPT-4 performs this task effectively, smaller open-source models may encounter challenges at this step. Additionally, users must configure certain hyperparameters for the system, including the \textit{Maximum of Expansion} and the \textit{Number of Branches}. The optimal values for these parameters may vary depending on the specific task.

\bibliography{master}

\appendix

\section{Pseudo Code}
\label{sec:appendix_a}
The pseudo code of our framework is demonstrated in this section.

After initializing a root agent with the question, multiple LLM calls are made in 'root.action()' to get the solution, validation and assessment of this root agent. The score and confidence are extracted from assessment. If this is a terminal agent, LLM is called again to decide whether its solution can pass the evaluation. All the information above including solution, validation, assessement, score, confidence and pass (bool) is saved in the memory of this agent and is passed to its child agent for informaion. In this pseudo code, only pass, solution and score are explicitly expressed for simplicity but all of them are obtained in our full code.

Then the node with the highest UCT is selected by the function 'select\_with\_uct(root)'. This function will go through all the agents under root agent, which is the entire reasoning tree to select the one with the highest UCT for expansion. This procedure is called Selection which will be done $i$ (Maximum of Expansion) times.

When an agent is selected, the algorithm will generate a few child agent. Every generation is done by 'agent.expand()' which will initialize a new agent and generate its own solution, validation, assessment, etc. like above. The only difference is that the root agent only has the question in its prompt but these child agent has all the context of the agents on the path from the root to itself in its prompt (its parent, parent of parent, etc.). This creatation of child agent will be repeated $j$ (Number of Branches) times which is called Expansion.

Whenever a terminal agent is generated, the evaluation as mentioned above is conducted. If it fails the evaluation (the variable pass is False), a procedure of Backpropagation is triggered. The reward of the agents on the path to this terminal agent will be updated with the reward of it.

\begin{algorithm}[tb]
\caption{MASTER algorithm}
\label{alg:algorithm}
\textbf{Input}: Question\\
\textbf{Parameter}: Expansion Maximum, Branches Number\\
\textbf{Output}: Solution
\begin{algorithmic}[1]
\STATE Initialize root with Question and Generate its Context
\STATE $root$ ← Agent($question$)
\STATE $pass, solution, sc$ ← $root$.action()
\FOR{$i$ ← 1, ..., Expansion Maximum}
\STATE \textsc{Selection}
\STATE $agent$ ← select\_with\_uct($root$)
\FOR{$j$ ← 1, ..., Branches Number}
\STATE \textsc{Expansion}
\STATE $pass, child\_solution, child\_score$ ← $agent$.expand()
\IF{$pass$ is True}
\STATE \textbf{return} solution
\ELSIF{$pass$ is False and $child$ is Terminal}
\STATE \textsc{Backpropagation}
\WHILE{$agent$ is not None}
\STATE $agent$.value ← update($child\_score$)
\STATE $agent$.visits ← $agent$.visits + 1
\STATE $agent$ ← $agent$.parent
\ENDWHILE
\ENDIF
\ENDFOR
\ENDFOR
\end{algorithmic}
\end{algorithm}

\section{Deduction of UCT Formula}
\label{sec:appendix_b}
In the methodology section, we proposed a modification to the UCT formula. Here, we provide the derivation of the original UCT formula.

According to Hoeffding's Inequality, for a random variable $r$ such that $r\in\left[a,b\right]$, if $r_1, r_2, ... r_n$ are independent and $r_i\in\left[a,b\right]$ almost surely with $a_i<b_i$ for all $i$, then:
\begin{align} 
    \label{eq:appendix1}
    P\left(\frac{1}{n_i}\sum_{n=1}^{n_i}(r_n - E[r_n]) \geq \varepsilon \right) \nonumber \\ \leq \exp \left(\frac{-2{n_i}^2 \varepsilon^2}{\sum_{n=1}^{n_i} (b_n - a_n)^2} \right)
\end{align}

The following term could be rewritten as:
\begin{align}
  \label{eq:appendix2}
  \frac{1}{n_i} \sum_{n=1}^{n_i}(r_n - E[r_n]) = \frac{1}{n_i} \sum_{n=1}^{n_i} r_n - \frac{1}{n_i} \sum_{n=1}^{n_i} E[r_n] \nonumber\\ = \frac{1}{n_i} \sum_{n=1}^{n_i} r_n - E[r_n]
\end{align}
In Eqn \ref{eq:appendix2}, the second term is the mean of the random variable $r_n$ which is noted as $q_i$. The first term is the sample mean or the estimation of $q_i$ which is noted as $Q_i$. If $r_n$ as reward is always between 0 and 1, $\sum_{i=1}^{n_i} (b_n - a_n)^2=\sum_{n=1}^{n_i} 1=n_i$. Then Eqn \ref{eq:appendix1} is:
\begin{equation}
  \label{eq:appendix3}
  P(Q_i-q_i\geq \varepsilon)\leq \exp\left(-2{n_i}^2 \varepsilon^2\right)
\end{equation}
Consider the symmetry:
\begin{equation}
  \label{eq:appendix4}
  P(Q_i-q_i\leq -\varepsilon)\leq \exp\left(-2{n_i}^2 \varepsilon^2\right)
\end{equation}
Combine Eqn \ref{eq:appendix3} and Eqn \ref{eq:appendix4}:
\begin{equation}
  \label{eq:appendix5}
  P(-\varepsilon\leq Q_i-q_i\leq\varepsilon)\geq 1-2 \cdot \exp(-2{n_i}^2 \varepsilon^2)
\end{equation}
Let $N_i=\exp{(2{n_i}\varepsilon^2)}$, then
\begin{equation}
  \label{eq:appendix6}
  \varepsilon=\sqrt{\frac{\ln{\left(N_i\right)}}{2n_i}}
\end{equation}
When we consider the upper bound of the reward:
\begin{equation}
  \label{eq:appendix7}
  Q_i-q_i\geq-\varepsilon
\end{equation}
\begin{equation}
  \label{eq:appendix8}
  q_i\le Q_i+\sqrt{\frac{\ln{\left(N_i\right)}}{2n_i}}
\end{equation}
Eqn \ref{eq:appendix8} is the normal UCT formula mentioned in our paper.

\section{Case Study}
\label{sec:appendix_c}
\begin{figure*}[ht]
    \centering
    \includegraphics[width=\linewidth]{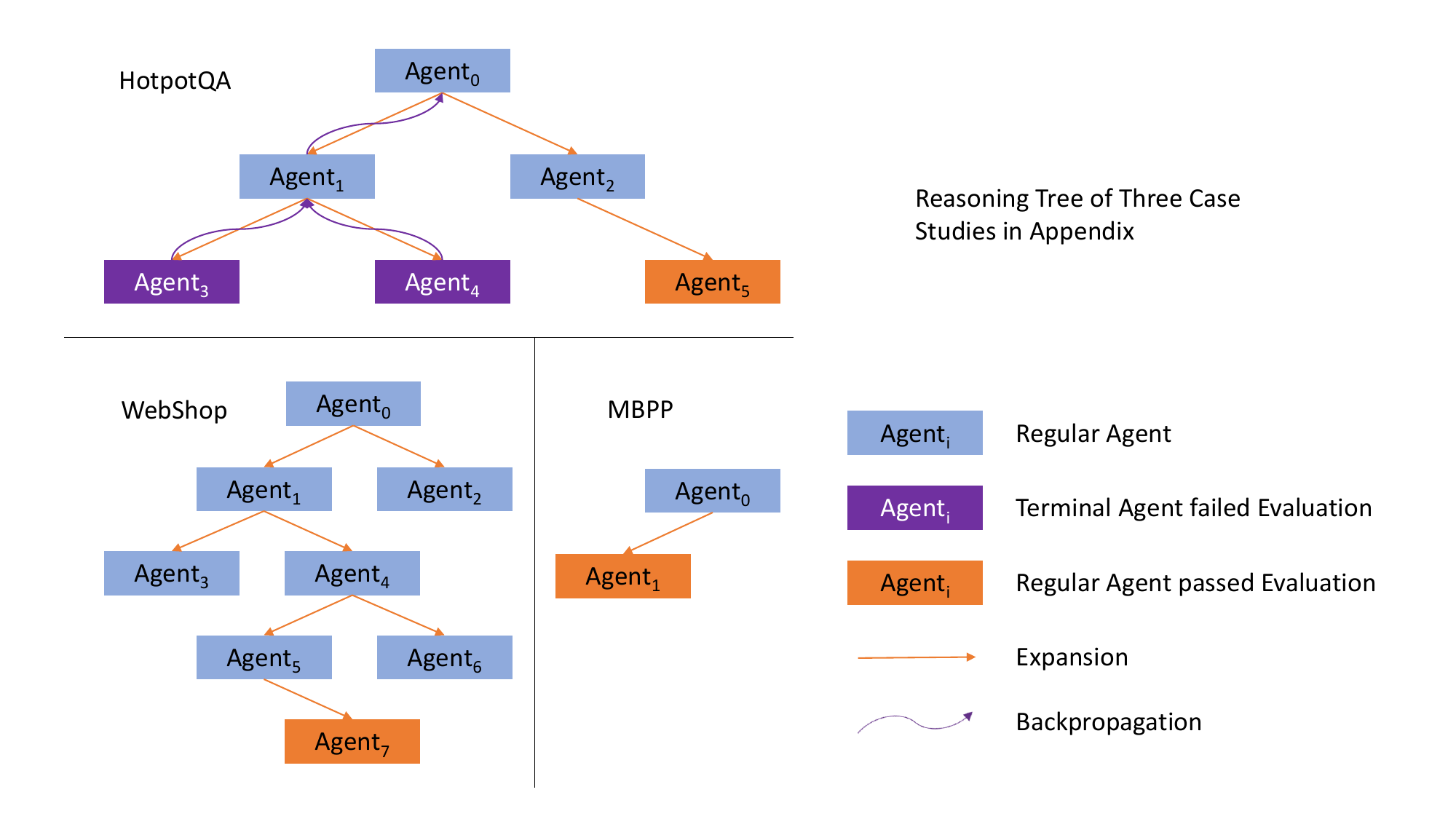}
    \caption{Reasoning Tree of case studies on three datasets.}
    \label{fig:case_study}
\end{figure*}

\subsection{HotpotQA}
Question: Dawn French's Girls Who Do Comedy included which American comedienne and television host?

\textbf{Agent$_0$}

Thought: The question asks for the name of an American comedienne and television host who was included in Dawn French's Girls Who Do Comedy. To answer this question, I need to find out which American female comedians were featured on that series.

Action: Search[Dawn French's Girls Who Do Comedy]

Observation: Dawn French's Girls Who Do Comedy is an interview series shown on BBC Four... Whoopi Goldberg, Catherine Tate, Kathy Burke, Julie Walters, Victoria Wood and Joan Rivers...

Validation: The latest 'Thought' is looking for information about Dawn French's Girls Who Do Comedy which aligns with the 'Question' that mentioned the interview series. The latest 'Action' searched for Dawn French's Girls Who Do Comedy as instructed by the latest 'Thought', so they are aligned.

Assessment: The trajectory has successfully found out comediennes included in Dawn French's Girls Who Do Comedy. However, the trajectory has not identify the American comedienn and television host. Thus the correctness score is 3. The confidence in this score is 9.

\textbf{Agent$_1$}

Thought: Whoopi Goldberg, Catherine Tate, Kathy Burke, Julie Walters, Victoria Wood and Joan Rivers were included in Dawn French's Girls Who Do Comedy. To answer the question, I need to search everyone of them for more information.

Action: Search[Whoopi Goldberg]

Observation: Caryn Elaine Johnson (born November 13, 1955), known professionally as Whoopi Goldberg, is an American actor, comedian, author, and television personality...

Validation: The latest 'Thought' is looking for information about Whoopi Goldberg, Catherine Tate, Kathy Burke, Julie Walters, Victoria Wood and Joan Rivers which aligns with the 'Observation' that they are included in Dawn French's Girls Who Do Comedy. The latest 'Action' searched for Whoopi Goldberg as instructed by the latest 'Thought', so they are aligned.

Assessment: This trajectory has enough information to answer the question since Whoopi Goldberg is an American comedienne who is included in Dawn French's Girls Who Do Comedy. Thus the correctness score is 8. The confidence in this score is 7.

\textbf{Agent$_2$}

Thought: The question asks for the name of an American comedienne in Dawn French's Girls Who Do Comedy. To answer the question, I need to search American comedienne.

Action: Search[American comedienne]

Observation: Katherine Beth Nolan (born January 28, 1987) is an American sports personality, comedienne, and television host on Fox Sports 1... Joan Alexandra Molinsky (June 8, 1933 – September 4, 2014), widely known as Joan Rivers, was an American comedian, actress, writer, producer, and television host...

Validation: The latest 'Thought' is looking for American comedienne which aligns with the 'Observation' that comediennes included in Dawn French's Girls Who Do Comedy are listed. The latest 'Action' searched for American comedienne as instructed by the latest 'Thought', so they are aligned.

Assessment: This trajectory searched American comedienne but did identify the one included in Dawn French's Girls Who Do Comedy. Thus the correctness score is 6. The confidence in this score is 9.

\textbf{Agent$_3$}

Thought: Whoopi Goldberg is an American comedienne who is included in Dawn French's Girls Who Do Comedy and all the information needed is found. I should answer the question now.

Action: Finish[Whoopi Goldberg]

Validation: The latest 'Thought' indicated to answer the question with Whoopi Goldberg which does not align with the 'Observation' because it is not mentioned that Whoopi Goldberg is a television host. The latest 'Action' finished the question with answer 'Whoopi Goldberg' as instructed by the latest 'Thought', so they are aligned.

Assessment: The answer is incorrect according to the validation because it is not mentioned that Whoopi Goldberg is a television host. Thus the correctness score is 2. The confidence in this score is 7.

Evaluation: False

\textbf{Agent$_4$}

Thought: Whoopi Goldberg is an American comedienne included in Dawn French's Girls Who Do Comedy and all the information needed is found. I should answer the question now.

Action: Finish[Whoopi Goldberg]

Validation: The latest 'Thought' answers the question with Whoopi Goldberg which does not align with the 'Observation' because it is not mentioned that Whoopi Goldberg is a television host. The latest 'Action' finished the question with answer 'Whoopi Goldberg' as instructed by the latest 'Thought', so they are aligned.

Assessment: The answer is incorrect according to the validation because Whoopi Goldberg is not a television host. Thus the correctness score is 2. The confidence in this score is 7.

Evaluation: False

\textbf{Agent$_5$}

Thought: Joan Rivers was an American comedian and television host who is included in Dawn French's Girls Who Do Comedy and all the information needed is found. I should answer the question now.

Action: Finish[Joan Rivers]

Validation: The latest 'Thought' indicated to answer the question with Joan Rivers which aligns with the 'Observation' because Joan Rivers is an American comedian and a television host who is included in Dawn French's Girls Who Do Comedy. The latest 'Action' Finished the question with answer 'Joan Rivers' as instructed by the latest 'Thought', so they are aligned.

Assessment: The answer is correct according to the validation because all the informaion is aligned. Thus the correctness score is 10. The confidence in this score is 9.

Evaluation: True

\subsection{WebShop}
Question: Instruction: Find me dual band streaming media players with quad core, and price lower than 350.00 dollars

\textbf{Agent$_0$}

Thought: The instruction requires finding a dual-band streaming media player with a quad-core processor and a price lower than \$350.00. Since there is a search bar available, the best course of action is to use it to search for products that match the criteria.

Action: search[dual band streaming media player with quad core]

Observation: Product index is B09LSKQF8C. Product description is Superbox S3 Pro Dual Band Wi-Fi 2.4Ghz 5Ghz Supports 6K Video. Product price is \$329.0... Available actions: has search bar: False, clickables: [b09lskqf8c, b09nmzs7hk, b00r249c5g, b07srs3mxp, b00ofqr8d2, b0871vqt3q, b085vfmxmz, b01lzwdnx6, b08qsnm69h, b08m5byqn9]

Validation: The common attributes include dual-band, streaming media player, quad-core, and price lower than \$350.00, while there are no optional attributes specified in the instruction.
The group of products to be discarded (Group 1) includes B09NMZS7HK, B00R249C5G, B07SRS3MXP, B00OFQR8D2, B0871VQT3Q, B085VFMXMZ, B01LZWDNX6, B08QSNM69H, and B08M5BYQN9 because they either do not match the common attributes of being a dual-band streaming media player with a quad-core processor or exceed the specified price limit.
The group of products to be considered (Group 2) includes B09LSKQF8C because it matches the common attributes of being a dual-band streaming media player with a quad-core processor and is priced below \$350.00.

Assessment: The product B09LSKQF8C meets all the specified requirements: it is a dual-band streaming media player with a quad-core processor and is priced below \$350.00. Thus, this product is a suitable choice and the task is close to completion, pending a final review of the product details and the purchase action. Thus the correctness score is 6. The confidence in this score is 9.

\textbf{Agent$_1$}

Thought: The product B09LSKQF8C seems to fit the requirements as it is a dual-band streaming media player and is priced below \$350.00. However, the product description does not mention if it has a quad-core processor. To ensure that it meets all the criteria, I need to click on the product index B09LSKQF8C to view more detailed information before making a purchase decision.

Action: click[b09lskqf8c]

Observation: Product description is Superbox S3 Pro Dual Band Wi-Fi 2.4Ghz 5Ghz Supports 6K Video. Price: \$329.0. Click corresponding options if needed. Click 'features' if you need more information. Click 'buy now' if you decide to purchase this product. Available actions: has search bar: False, clickables: [back to search, prev, description, features, reviews, buy now]

Validation: The common attributes include dual-band streaming media players, quad-core, and price lower than 350.00 dollars, while there are no optional attributes specified in the instruction. The product Superbox S3 Pro fits the common attributes as it is a dual-band streaming media player and is priced below \$350.00. However, the product description does not mention if it has a quad-core processor, which is a required common attribute.

Assessment: The latest 'Thought' indicates that the quad-core processor attribute is crucial and has not yet been confirmed for the Superbox S3 Pro. The 'Action' taken is to click on 'features' to verify this attribute, and the 'Observation' is pending, awaiting confirmation on whether the quad-core processor is present. If the feature is confirmed, the task is close to completion; otherwise, further search is required. Thus the correctness score is 7. The confidence in this score is 8.

\textbf{Agent$_2$}

\textbf{Agent$_2$} is similar to \textbf{Agent$_1$}, omitted due to space limitation.

\textbf{Agent$_3$}

Thought: The Superbox S3 Pro mentions that it supports dual-band Wi-Fi and 6K video, and the price is listed as \$329.00, which meets the price requirement. However, it is not confirmed whether the device has a quad-core processor. The details could be found in 'description' section. It is necessary to check the 'description' to find out the specified requirements before proceeding with the purchase.

Action: click[description]

Observation: The newest SuperBox S3 Pro is an advanced voice control IPTV box, with a brand new Bluetooth remote and built-in artificial intelligence system, you can easily speak to control your TV box. The box has 2GB of RAM and 32GB of storage, utilizes a quad-core ARM Cortex-A53 processor, supports Android 9.0 OS. The new model has a new flat rectangular shape of design with an additional led display and new interface, the SuperBox S3 Pro has also added more features. It adopts 2T2R WiFi technology makes our streaming device 60\% faster and more stable. In conclusion, the SuperBox S3 Pro is really the game-changer in streamer, it is by far the most powerful and most user-friendly TV Box on the market. Available actions: has search bar: False, clickables: [back to search, prev]

Validation: The common attributes include dual-band streaming media players, quad-core, and price lower than 350.00 dollars, while there are no optional attributes specified in the instruction. It also has a quad-core processor, which is a required common attribute.

Assessment: The latest 'Thought' indicates that the quad-core processor attribute is crucial and is not confirmed for the Superbox S3 Pro. The 'Action' taken is to click on 'description' to verify this attribute, as quad-core processor is present, the task is close to completion. Thus the correctness score is 7. The confidence in this score is 7.

\textbf{Agent$_4$}

Thought: The product description provided for the Superbox S3 Pro mentions that it supports dual-band Wi-Fi and 6K video, and the price is listed as \$329.00, which meets the price requirement. However, the description does not confirm whether the device has a quad-core processor. The 'features' section is likely to contain more detailed specifications about the product, including the processor details. It is necessary to check the 'features' to ensure that the product meets all the specified requirements before proceeding with the purchase.

Action: click[features]

Observation: Prev Built-in Voice Control System Android 9.0 OS Quad-core ARM Cortex-A53 Processor 2T2R( 2 Transmitter, 2 receivers) antenna and 2.4G/5G Dual Band WiFi 2GB DDR3 Memory 32GB eMMC Internal Storage Support 4K@60fps. Available actions: has search bar: False, clickables: [back to search, prev]

Validation: The common attributes include dual-band streaming media players, quad-core processor, and price lower than \$350.00 dollars. There are no optional attributes specified in the instruction.
The current state indicates that the product B09LSKQF8C, identified as the Superbox S3 Pro, has been clicked on and its features have been examined. The features confirm that the product has a quad-core ARM Cortex-A53 processor and supports 2.4G/5G dual-band Wi-Fi. The price of \$329.00 is also within the specified budget. Since the product meets all the common attributes specified in the instruction, it is the correct product to select.
The trajectory does not require reselection of any attributes, as there are no optional attributes to consider, and the product matches all the common attributes. Therefore, the next step should be to proceed with the purchase.

Assessment: The product B09LSKQF8C, Superbox S3 Pro, has been confirmed to have a quad-core ARM Cortex-A53 processor, supports dual-band Wi-Fi, and is priced at \$329.00, which meets all the specified requirements. The task is close to being solved as the only remaining step is to initiate the purchase by clicking 'buy now'. Thus the correctness score is 9. The confidence in this score is 8.

\textbf{Agent$_5$}

Thought: The features of the Superbox S3 Pro confirm that it has a quad-core ARM Cortex-A53 processor, supports 2.4G/5G dual-band WiFi, and is within the specified price range. Since it meets all the requirements outlined in the instruction, the next step is to go back to the previous page where the 'buy now' button is available and complete the purchase.

Action: click[prev]

Observation: Product description is Superbox S3 Pro Dual Band Wi-Fi 2.4Ghz 5Ghz Supports 6K Video. Price: \$329.0. Click corresponding options if needed. Click 'features' if you need more information. Click 'buy now' if you decide to purchase this product.Available actions: has search bar: False, clickables: [back to search, prev, description, features, reviews, buy now]

Validation: The common attributes include dual-band streaming media players with a quad-core processor and a price lower than \$350.00. There are no optional attributes specified in the instruction.
The current state shows that the Superbox S3 Pro meets all the common attribute requirements: it supports dual-band Wi-Fi, has a quad-core ARM Cortex-A53 processor, and is priced at \$329.00, which is below the \$350.00 threshold. Since there are no optional attributes to consider and the product matches all the common attributes, the next logical step is to proceed with the purchase.
The topic of the response is to remind the agent to complete the purchase.
Since the product meets all the necessary criteria and there are no optional attributes to adjust, the agent should click 'buy now' to finalize the purchase.

Assessment: The product Superbox S3 Pro has been confirmed to meet all the specified requirements, including the dual-band capability, quad-core ARM Cortex-A53 processor, and the price point of \$329.00. The next step is to click 'buy now' to proceed with the purchase, as there are no optional attributes to adjust and all common attributes are satisfied. Thus the correctness score is 9. The confidence in this score is 9.

\textbf{Agent$_6$} is similar to \textbf{Agent$_5$}, omitted due to space limitation.

\textbf{Agent$_7$}

Thought: The Superbox S3 Pro has been confirmed to meet all the requirements of the instruction, which include being a dual-band streaming media player with a quad-core processor and having a price lower than \$350.00. The 'buy now' button is available to complete the purchase.

Action: click[buy now]

Observation: Product description is Superbox S3 Pro Dual Band Wi-Fi 2.4Ghz 5Ghz Supports 6K Video. Price: \$329.0. Click corresponding options if needed. Click 'features' if you need more information. Click 'buy now' if you decide to purchase this product.Available actions: has search bar: False, clickables: [back to search, prev, description, features, reviews, buy now]

Validation: The common attributes include dual-band streaming media players with a quad-core processor and a price lower than \$350.00. There are no optional attributes specified in the instruction.
The current state shows that the Superbox S3 Pro meets all the common attribute requirements and there are no optional attributes to consider.

Assessment: The product Superbox S3 Pro has been confirmed to meet all the specified requirements, including the dual-band capability, quad-core ARM Cortex-A53 processor, and the price point of \$329.00. Thus the correctness score is 10, The confidence in this score is 10.

Evaluation: True

\subsection{MBPP}
Qustion: Write a function to count all the distinct pairs having a difference of k in any array.

Test cases: 

assert countpairs([1, 5, 3, 4, 2], 5, 3) == 2

assert countpairs([8, 12, 16, 4, 0, 20], 6, 4) == 5

assert countpairs([2, 4, 1, 3, 4], 5, 2) == 3

\textbf{Agent$_0$}

Thought: To solve this problem, I need to find all the unique pairs (a, b) in the array such that the absolute difference between a and b is k. To do this efficiently, I can use a set to keep track of the elements in the array for constant-time lookups. Then, I can iterate through the array and for each element, I can check if the element plus k or the element minus k exists in the set. If it does, it constitutes a distinct pair with the required difference. I should also ensure that I count each pair only once.

Action:

def countpairs(arr, n, k):

    unielements = set(arr)
    
    count = 0
    
    for number in unielements:
    
        if (number + k) in unielements:
        
            count += 1
            
        if (number - k) in unielements:
        
            count += 1
    
    return count // 2

Observation: 

Tests passed:

assert countpairs([1, 5, 3, 4, 2], 5, 3) == 2

assert countpairs([8, 12, 16, 4, 0, 20], 6, 4) == 5

Tests failed:

assert countpairs([2, 4, 1, 3, 4], 5, 2) == 3 while output: 2

Passed 2 out of 3 tests. Failed 1 out of 3 tests.

Validation: The code did not pass the last test case. The issue lies in the logic used to count the pairs. The current implementation incorrectly counts a pair twice if both `number + k` and `number - k` exist in the set of unique elements. However, this approach does not account for duplicate elements in the array, which can lead to incorrect pair counts.

Assessment: The code is almost correct but fails one test case due to a logical error in counting pairs. Thus the correctness score is 7. The confidence in this score is 10.

\textbf{Agent$_1$}

Thought: The previous code does not correctly handle duplicates where the difference between numbers should be counted multiple times if there are multiple occurrences of a number. I should use a dictionary to keep track of the occurrences of each number as we iterate through the array.

Action:

def countpairs(arr, n, k):

    count = 0
    
    seen = \{\}
    
    for num in arr:
        
        if num + k in seen:
        
            count += seen[num + k]
        
        if num - k in seen:
        
            count += seen[num - k]

        if num in seen:
        
            seen[num] += 1
            
        else:
        
            seen[num] = 1
            
    return count

Observation:

Tests passed:

assert countpairs([1, 5, 3, 4, 2], 5, 3) == 2

assert countpairs([8, 12, 16, 4, 0, 20], 6, 4) == 5

assert countpairs([2, 4, 1, 3, 4], 5, 2) == 3

Tests failed:

Passed 3 out of 3 tests. Failed 0 out of 3 tests.

Validation: Since the code has passed all the tests and there are no observed errors or failures, I suggest the evaluator to pass the code.

Assessment: The code is correct and passes all the test cases. Thus the correctness score is 10. The confidence in this score is 10.

Evaluation: True

\end{document}